\documentclass[letterpaper]{article}
\PassOptionsToPackage{table}{xcolor}
\usepackage[preprint]{aaai2027}
\usepackage[hyphens]{url}
\usepackage{graphicx}
\usepackage{natbib}
\usepackage{caption}
\usepackage{amsmath}
\usepackage{amssymb}
\usepackage{mathtools}
\usepackage{amsthm}
\usepackage{booktabs}
\usepackage{multirow}
\usepackage{array}
\usepackage{makecell}
\usepackage{algorithm}
\usepackage{algpseudocode}
\usepackage{microtype}
\makeatletter
\def\r@Distortion{{S1}{2}}
\def\r@Black{{S2}{3}}
\@namedef{r@alg:spcame}{{S1}{1}}
\@namedef{r@tab:ours_multishape_elastic_shapeaware}{{S1}{2}}
\@namedef{r@fig:geometric_qualitative}{{S3}{4}}
\@namedef{r@tab:jnd_ablation_noclip}{{S2}{3}}
\def\r@JND{{S4}{4}}
\makeatother

\graphicspath{{Fig/}}
\newcolumntype{C}[1]{>{\centering\arraybackslash}m{#1}}

\theoremstyle{plain}

\theoremstyle{definition}

\theoremstyle{remark}

\title{CASIAL: Geometric Distortion Robust Image Watermarking}
\author{
    Yupeng Qiu\textsuperscript{\rm 1},
    Han Fang\textsuperscript{\rm 2}\corresponding,
    Ee-Chien Chang\textsuperscript{\rm 1}
}
\affiliations{
    \textsuperscript{\rm 1}School of Computing, National University of Singapore, Singapore\\
    \textsuperscript{\rm 2}School of Cyber Science and Technology, University of Science and Technology of China, Hefei, Anhui, China\\
    qiu\_yupeng@u.nus.edu, fanghan@ustc.edu.cn, changec@comp.nus.edu.sg
}
\begin{document}

\maketitle

\begin{abstract}

Deep learning–based watermarking has shown strong robustness against non-geometric distortions, yet its performance under geometric transformations remains limited. Such transformations induce two fundamental failure modes: region removal, such as cropping or masking, which eliminates the information carried by removed pixels, and desynchronization, such as scaling or rotation, which misaligns pixel positions and disrupts decoding. We argue that achieving geometric robustness requires two essential properties: (1) global spread of the watermark message, ensuring resilience even when large regions are removed, and (2) geometry-invariant representations, enabling decoding to remain synchronized despite spatial transformations. Building on these insights, we propose CASIAL, a geometric distortion–robust watermarking framework with \textbf{c}over image-\textbf{a}ware message \textbf{s}preading (CAS) strategy and \textbf{i}nvariance \textbf{a}lignment \textbf{l}earning (IAL) module. CAS tightly couples watermark bits with cover image features and distributes them adaptively across the entire image, enhancing per-pixel information capacity and robustness to region removal. IAL leverages spatial attention to capture cross-pixel dependencies and align perturbed features into a shared geometry-invariant representation space, mitigating failures due to desynchronization. 
Across six challenging geometric transformations, CASIAL achieves substantially stronger robustness than eleven prior baselines while preserving high visual quality.
It also maintains competitive performance under six signal distortions and four photometric transformations. Notably, although trained only with white-box distortions, CASIAL also exhibits strong transfer robustness to unseen black-box distortions. Comprehensive experiments demonstrate the broad robustness and superior visual quality of our method.
\end{abstract}

\section{Introduction}
\label{sec:intro}
Digital watermarking has been widely studied across digital media \citep{hartung1999multimedia,petitcolas1999information} and is commonly used for copyright protection \citep{cox1997secure,barni1998dct}, authorship verification \citep{melman2020authorship,sharma2024review}, and provenance tracing \citep{fernandez2023stable, yang2024gaussian, li2025gaussmarker}. In a typical pipeline, a hidden message is encoded into an image and later decoded to verify ownership or origin. A practical watermarking system must balance two objectives: invisibility (minimal perceptual impact) and robustness (reliable recovery under distortions).
\begin{figure}[!t]
    \centering
    \includegraphics[width=\columnwidth]{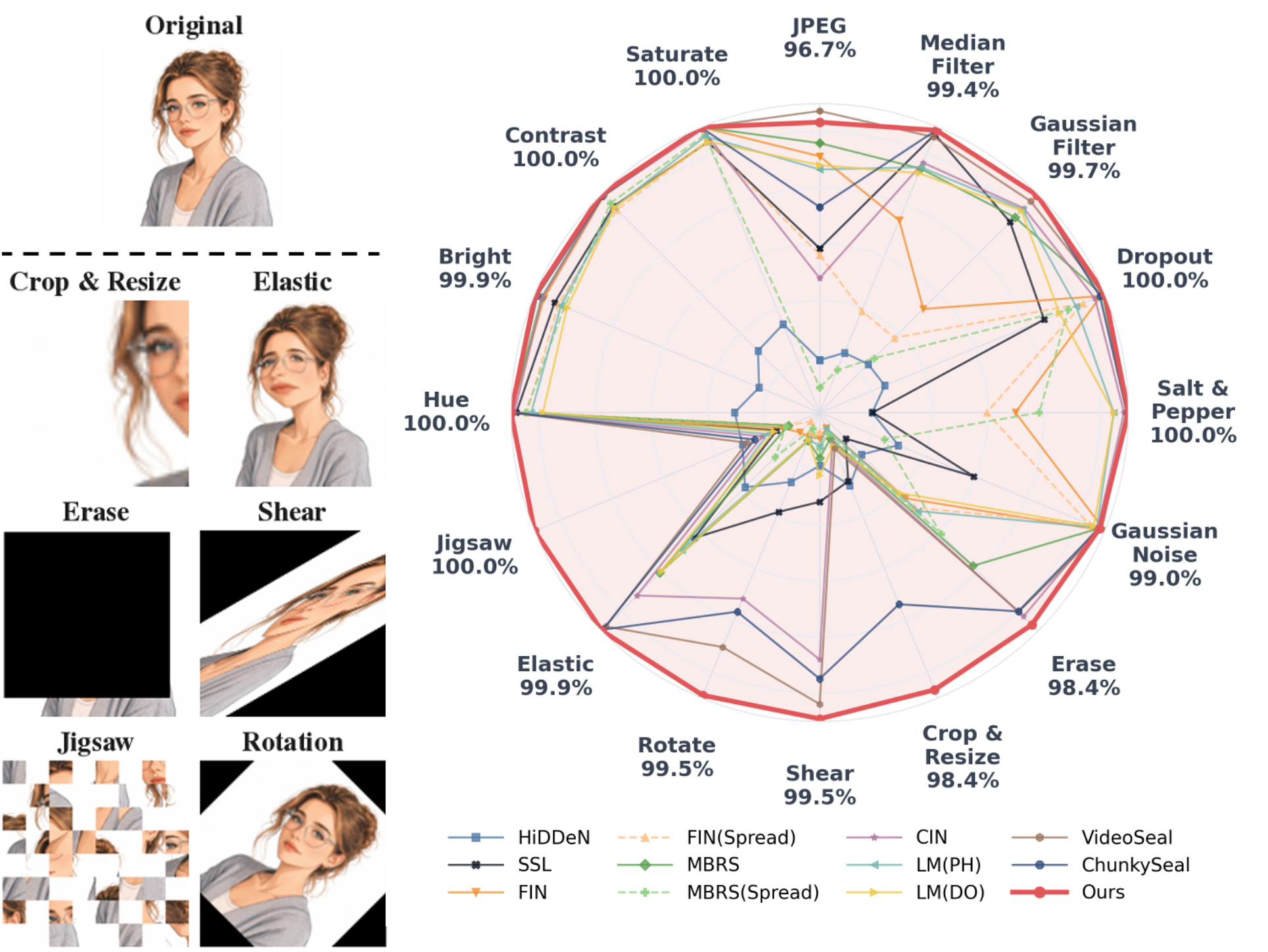}
    \caption{Robustness comparison under different distortions.}
    \label{RadorFig}
    \vspace{-0.2cm}
\end{figure}

Recent deep learning-based watermarking methods \citep{zhu2018hidden, zhang2021towards,jia2021mbrs,fernandez2022watermarking, ma2022towards, fang2023flow,fernandez2024video,qiu2025lightweight,petrov2025we} usually improve robustness by inserting differentiable distortion layers during training, so the encoder–decoder learns distortion-tolerant representations. This strategy has proved effective for non-geometric distortions such as compression, blur, and additive noise. Robustness to geometric transformations remains limited.

Geometric transformations are qualitatively different from non-geometric distortions because they alter spatial correspondence rather than merely perturbing pixel values. We summarize their effects into two primary failure modes: \textbf{region removal} (e.g., cropping or masking), which discards subsets of pixels and irreversibly erases the watermark evidence they carry; and \textbf{desynchronization} (e.g., scaling or rotation), which changes spatial alignment so that a decoder reading at fixed locations becomes mismatched and fails even when the signal is still present.

These failure modes reveal two missing capabilities in current deep watermarking pipelines. First, without an effective mechanism to distribute message evidence globally, extraction becomes brittle when large regions are removed. Second, conventional CNN backbones, biased toward local receptive fields and fixed spatial layouts, struggle to maintain synchronization or form geometry-invariant representations under spatial warps. This motivates a key insight: geometric robustness requires two complementary properties: (i) global message spread to survive region removal, and (ii) geometry-invariant representations to remain synchronized under spatial transformations.

To realize these properties, we propose \textbf{CASIAL}, a geometric distortion–robust watermarking framework with two main components: \textbf{c}over image-\textbf{a}ware message \textbf{s}preading (CAS) strategy and \textbf{i}nvariance \textbf{a}lignment \textbf{l}earning (IAL) module. CAS effectively spreads the watermark message over the entire image space, increasing the per-pixel information capacity so that reliable decoding remains possible even when large regions are removed (e.g., heavy cropping or masking).
IAL leverages spatial attention \citep{vaswani2017attention} to capture cross-pixel dependencies and align perturbed features into a shared geometry-invariant representation space, thereby preserving synchronization under spatial transformations.

Specifically, CAS does not treat message bits as independent symbols to be injected at isolated locations. Instead, it uses each bit to modulate cover-conditioned candidate features, so that the resulting message-dependent features are derived from the cover image and their evidence is naturally distributed across spatial locations. This cover-aware spreading increases the amount of recoverable evidence remaining after aggressive cropping or masking. Meanwhile, IAL integrates spatial attention into both the encoder and decoder, enabling the model to capture long-range dependencies and reweight spatial regions adaptively, downweighting corrupted or missing areas while upweighting reliable content. Together, CAS and IAL realize the two essential operations for geometric robustness, allowing CASIAL to withstand both region removal and desynchronization while preserving invisibility and fidelity.

In summary, our contributions are as follows:
\begin{itemize}
\item We analyze why geometric transformations remain challenging for deep watermarking and identify two fundamental failure modes, region removal and desynchronization, which motivate two required properties for geometric robustness: global message spread and geometry-invariant representations.

\item We propose CASIAL, a geometry-robust watermarking framework that realizes these properties through two components: (i) cover image-aware message spreading (CAS), which couples watermark bits with cover features and distributes them across the image; and (ii) invariance alignment learning (IAL), which uses spatial attention to align distorted features in a geometry-invariant representation space.

\item Extensive experiments against eleven state-of-the-art baselines show that CASIAL achieves superior robustness to six geometric transformations, maintains strong performance under six signal and four photometric distortions, generalizes to five black-box distortions and diverse image shapes, and preserves high visual quality.

\end{itemize}

\section{Related Work}
\label{sec:related}
\subsection{Deep learning-based watermarking}
Deep learning-based watermarking has largely replaced hand-crafted transform-domain methods \citep{huang2001dwt, ahmidi2004novel, ingemar2008digital} by learning distortion-tolerant representations directly from data.
Existing methods can be broadly categorized into three lines. The most common is the encoder-noise-layer-decoder (END) framework \citep{zhu2018hidden, zhang2021towards, jia2021mbrs, fernandez2024video, qiu2025lightweight, petrov2025we}, where an encoder embeds a message into an image and a decoder recovers it from the watermarked image. A second line uses invertible architectures, such as CIN \citep{ma2022towards} and FIN \citep{fang2023flow}, to perform encoding and decoding through the same model in opposite directions. A third line formulates encoding as image optimization; for example, SSL \citep{fernandez2022watermarking, qiurevisiting} optimizes an image against a pretrained decoder so that it carries a recoverable watermark.
Across these approaches, robustness is typically improved by simulating distortions with noise layers during training. Existing studies have developed differentiable approximations for non-differentiable distortions \citep{zhang2021towards}, mixed simulated and real JPEG compression \citep{jia2021mbrs}, and invertible noise layers for black-box distortions \citep{fang2023flow}. Other methods model physical capture processes to improve robustness to camera capture and screen shooting \citep{tancik2020stegastamp, fang2022pimog}. Despite these advances, geometric robustness remains a major challenge: even with geometric noise layers, existing methods still suffer substantial performance degradation under geometric transformations, as shown in Fig.~\ref{RadorFig}.


\subsection{Transformers and self-attention}
The design goals of Invariance Alignment Learning (IAL) are well suited to Transformer-based architectures. IAL requires long-range dependency modeling and global information exchange across spatial regions, capabilities naturally supported by self-attention~\citep{vaswani2017attention}. Vision Transformers have demonstrated the effectiveness of global token interactions for image representation~\citep{dosovitskiy2020image,touvron2021training}, while hierarchical variants such as Swin improve computational efficiency and multi-scale modeling through windowed attention~\citep{liu2021swin}. Transformer architectures have also been widely adopted in image restoration and enhancement, where aggregating non-local context is important for handling spatially heterogeneous degradations~\citep{zamir2022restormer,wang2022uformer}. Motivated by these advances, we implement IAL with spatial attention to capture long-range dependencies and adaptively reweight spatial regions. By aggregating image-dependent context across the feature map, IAL can suppress unreliable regions and emphasize informative ones, facilitating the alignment of perturbed features into a geometry-invariant representation space.

\begin{figure*}[!t]
\centering
\includegraphics[width=0.85\textwidth]{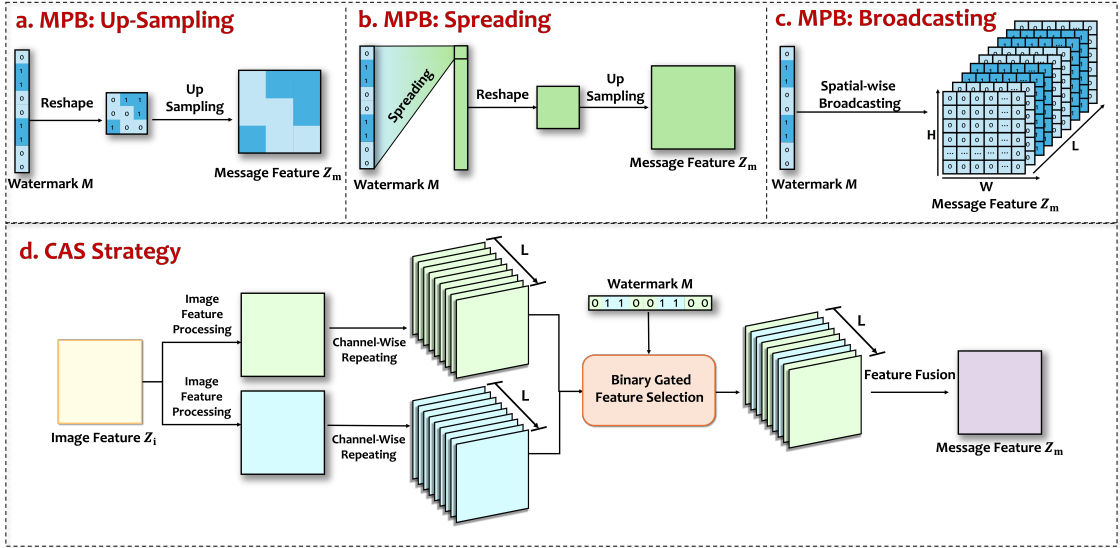}
\caption[Representative MPBs and CAS strategy]{Representative message processing blocks (MPBs) in prior deep watermarking methods and the proposed CAS strategy.
(a) Up-sampling MPB directly reshapes and upsamples the message, which is simple but tends to produce localized evidence.
(b) Spreading MPB first expands the message with a linear projection before reshaping and upsampling, increasing spatial coverage while still generating message features independently of the cover image.
(c) Broadcasting MPB repeats the message along spatial dimensions and concatenates it with image features, providing dense conditioning but imposing redundant message information at every location.
(d) CAS generates bit-dependent candidate features from the cover-image feature \(Z_{\mathrm{i}}\), uses a binary selector controlled by the message bits, and fuses the selected features into a cover-aware message feature \(Z_{\mathrm{m}}\).}
\label{MessageProcessBlockFig}
\end{figure*}

\subsection{Perceptual masking and JND attenuation}
Just-noticeable-difference (JND) models estimate the local distortion threshold below which changes are difficult for the human visual system to perceive \citep{chou1995perceptually,wu2017enhanced}.
Classical JND models commonly rely on luminance masking and contrast masking to describe spatially varying visual sensitivity.
Recent watermarking systems also use perceptual masks to guide invisible watermark encoding.
WAM \citep{sander2025watermark} uses a hand-crafted per-pixel perceptual map to modulate watermark intensity.
PixelSeal \citep{souvcek2025pixel} also adopts JND-based attenuation as part of its high-resolution adaptation to improve imperceptibility.

\begin{figure*}[!t]
\centering
\includegraphics[width=0.90\textwidth]{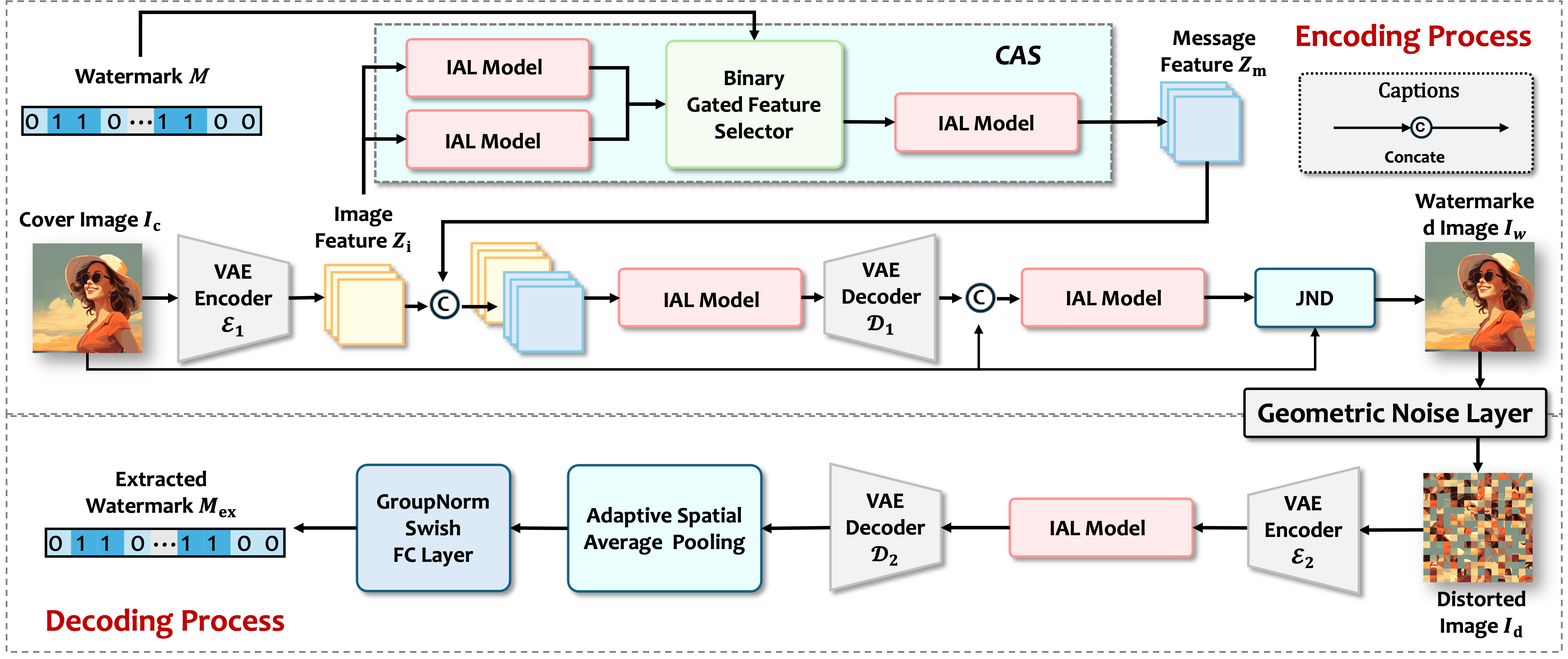}
\caption[Overview of CASIAL]{Overview of CASIAL. In the encoder, a VAE produces cover-image features, CAS produces a cover-aware message feature, IAL fuses the two, and a VAE decoder generates the raw watermarked image. JND-based residual attenuation then produces the final watermarked image. During training, a noise layer applies various distortions. In the decoder, a second VAE produces features from the distorted image, the IAL block refines them, and a VAE decoder with a linear head decodes the encoded message.}
\label{FramewrokFig}
\end{figure*}

\section{Motivation and Methods}
\subsection{Motivation}
\label{motivation:analysis-on-MPBs}
Current watermarking frameworks cannot gain geometric robustness from noise layers alone, 
because geometric attacks expose two missing capabilities that directly match the two failure modes: 
(i) how to spread message evidence globally to survive region removal, 
and (ii) how to adaptively aggregate spatial evidence to remain synchronized under desynchronization.\\
\textbf{Message spreading.}
Many deep watermarking frameworks employ a message-processing block (MPB) to transform a binary message into a spatial feature map before fusing it with image features. Figure~\ref{MessageProcessBlockFig} summarizes three representative designs.

The first design directly reshapes the message and progressively upsamples it using transposed convolutions~\citep{jia2021mbrs,fang2023flow,qiu2025lightweight}, as shown in Fig.~\ref{MessageProcessBlockFig}(a). Although lightweight and easy to integrate, this design can concentrate message evidence in a limited set of spatial patterns, making decoding vulnerable when the corresponding regions are cropped or erased.

The second design first projects the message into a higher-dimensional representation and then reshapes and upsamples it into a spatial feature map~\citep{jia2021mbrs,ma2022towards,fang2023flow}, as shown in Fig.~\ref{MessageProcessBlockFig}(b). The projection stage improves spatial coverage. However, the message feature is still generated independently of the cover image, which may cause a mismatch during fusion and lead to a less favorable trade-off between robustness and visual quality.

The third design broadcasts the message across the spatial dimensions and concatenates the resulting message map with image features~\citep{zhu2018hidden,fernandez2024video,petrov2025we}, as shown in Fig.~\ref{MessageProcessBlockFig}(c). This exposes the full message to every spatial location and therefore provides broad spatial coverage. However, repeatedly injecting the same cover-independent message representation can introduce redundant perturbations and increase the unnecessary visual burden of embedding.

These observations suggest that geometric robustness requires more than broad spatial coverage: message features should adapt to the cover image. CAS addresses this by deriving candidate features from the cover representation \(Z_{\mathrm{i}}\), selecting message-dependent candidates according to the embedded bits, and combining them into a cover-aware message representation \(Z_{\mathrm{m}}\), as illustrated in Fig.~\ref{MessageProcessBlockFig}(d).\\
\textbf{Spatial aggregation.} 
geometric transformations such as scaling or rotation mainly cause misalignment rather than erasing the signal, requiring the decoder to integrate displaced evidence and downweight unreliable regions. However, conventional CNN-style backbones\citep{zhu2018hidden, zhang2021towards,jia2021mbrs,fernandez2022watermarking, ma2022towards, fang2023flow,fernandez2024video,qiu2025lightweight,petrov2025we} with fixed kernels/strides lack an explicit mechanism to model long-range dependencies or reweight spatial regions adaptively, making recovery fragile under desynchronization.

These limitations motivate two requirements for geometric robustness: global message spreading to address region removal, and adaptive spatial aggregation/alignment to address desynchronization.

\subsection{Methods}

Our goal is to improve robustness to geometric transformations by addressing region removal and desynchronization explicitly. As illustrated in Fig.~\ref{FramewrokFig}, CASIAL follows the encoder-noise-layer-decoder framework and uses variational autoencoders (VAEs) \citep{rombach2022high} for feature encoding and fusion in latent space.\\
\textbf{Encoder.}
Given a cover image $I_{\mathrm{c}}\in\mathbb{R}^{3\times H\times W}$, the VAE encoder $\mathcal{E}_{1}$ produces a latent cover feature
\[
Z_{\mathrm{i}}=\mathcal{E}_{1}(I_{\mathrm{c}})\in\mathbb{R}^{C\times \tfrac{H}{4}\times \tfrac{W}{4}}.
\]
Let $M\in\{0,1\}^{L}$ denote a binary watermark of length $L$. The CAS block takes $(Z_{\mathrm{i}}, M)$ and outputs a message feature $Z_{\mathrm{m}}$. At a high level it derives $Z_{\mathrm{m}}$ directly from the cover image feature $Z_{\mathrm{i}}$ under bit controlled selection. The encoder concatenates $Z_{\mathrm{i}}$ and $Z_{\mathrm{m}}$, refines them with an IAL block,
\[
\tilde{Z}=\mathrm{IAL}\!\left(Z_{\mathrm{i}}\oplus Z_{\mathrm{m}}\right),
\]
where $\oplus$ denotes channel concatenation, and then maps the result back to image space with the VAE decoder $\mathcal{D}_{1}$.
A final IAL block combines the decoded feature with the cover image to produce a raw watermarked image:
\[
\tilde{I}_{\mathrm{w}} =
\mathrm{IAL}\!\left(\mathcal{D}_{1}(\tilde{Z})\oplus I_{\mathrm{c}}\right).
\]
We then apply JND-based residual attenuation to obtain the final watermarked image:
\[
I_{\mathrm{w}} =
I_{\mathrm{c}} +
{\mathrm{JND}}(I_{\mathrm{c}})
\odot
(\tilde{I}_{\mathrm{w}} - I_{\mathrm{c}}).
\]
Here, \({\mathrm{JND}}(I_{\mathrm{c}})\) denotes a perceptual visibility mask computed from the cover image.
This forward-path attenuation places stronger residuals in perceptually tolerant regions and suppresses them in visually sensitive regions.\\
\begin{figure*}[!t]
\centering
\includegraphics[width=0.85\textwidth]{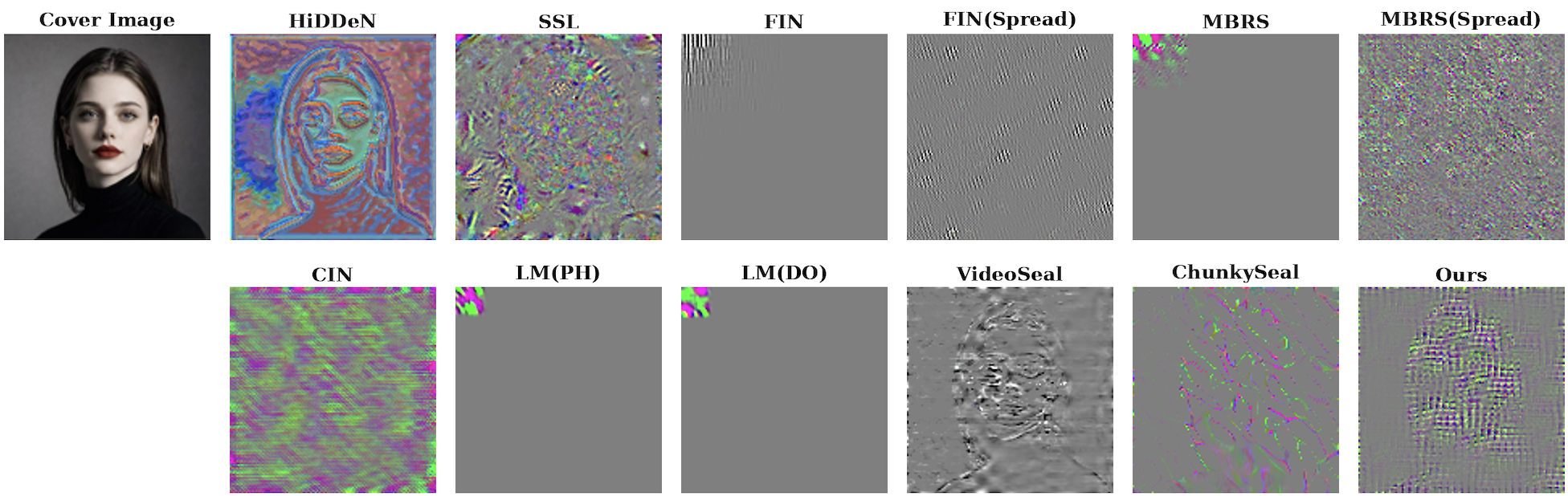}
\caption[Message spreading and cover-image coupling]{Visualization of message spreading and cover-image coupling. For each method and cover image, we encode (i) an all-zero message and (ii) the same message with only the first bit flipped, and then show the absolute difference between the two watermarked images. Image-wide residuals indicate broader spreading, while residual patterns that vary with the cover image indicate stronger content coupling.}
\label{InfoLoss}
\end{figure*}
\textbf{Noise layer.}
During training, the noise layer $\mathcal{N}$ applies simulated geometric transformations to the watermarked image to produce a distorted image 
\[
I_{\mathrm{d}}=\mathcal{N}(I_{\mathrm{w}}).
\]
\\
\textbf{Decoder.}
The decoder receives the distorted image $I_{\mathrm{d}}$. Another VAE encoder produces a latent feature
\[
Z'_{\mathrm{i}}=\mathcal{E}_{2}(I_{\mathrm{d}}),
\]
The features are refined by IAL, which suppresses corrupted regions and emphasizes reliable content. The VAE decoder $\mathcal{D}_{2}$, adaptive spatial average pooling, and a linear layer then recover the watermark $M_{\mathrm{ex}}$.

Overall, CAS enables cover-aware watermark spreading, IAL provides robust feature aggregation under spatial misalignment, and the noise layer promotes distortion-invariant learning.

\subsection{CAS Strategy}
Fig.~\ref{MessageProcessBlockFig}(d) illustrates the CAS strategy, while Algorithm~\ref{alg:spcame} in the Supplementary Material details its procedure.\\
\textbf{Binary candidate feature generation}
CAS takes latent cover-image feature $Z_{\mathrm{i}}$ and binary message $M$ as input. It first generates two candidate features from $Z_{\mathrm{i}}$ through two separate IAL blocks:
\[
F^{(0)}=\mathrm{IAL}_{0}(Z_{\mathrm{i}}), \qquad  F^{(1)}=\mathrm{IAL}_{1}(Z_{\mathrm{i}}),
\]
where $F^{(0)},F^{(1)}\in\mathbb{R}^{C\times \tfrac{H}{4}\times \tfrac{W}{4}}$ denote the candidate features corresponding to bit 0 and bit 1, respectively.
\\
\textbf{Binary gated feature selection}
The binary gated feature selector (BGFS) is a parameter-free selector that treats each bit \(m_{k}\in\{0,1\}\) as a control signal. Given candidate features \(F^{(0)}\) and \(F^{(1)}\) and message \(M=[m_{1},\ldots,m_{L}]\), we select for each bit $m_k$ a candidate feature as
\[
S_{k} \;=\; (1 - m_{k})\,F^{(0)} \;+\; m_{k}\,F^{(1)} \ .
\]
Concatenating \(\{S_{k}\}_{k=1}^{L}\) along the channel dimension gives
\[
\tilde{S} \;=\; \operatorname{Concat}_{\text{channel}}\!\big(S_{1},\ldots,S_{L}\big)
\;\in\; \mathbb{R}^{(L \cdot C) \times \tfrac{H}{4} \times \tfrac{W}{4}} \, .
\]
\textbf{Attention-based message feature fusion}
An IAL block fuses the concatenated channels in \(\tilde{S}\) and produces the final message feature
\[
Z_{\mathrm{m}} \;=\; \mathrm{IAL}_{\mathrm{2}}(\tilde{S}) \;\in\; \mathbb{R}^{C\times \tfrac{H}{4}\times \tfrac{W}{4}} \, .
\]
Through selection, CAS uses an entire candidate feature to represent bit zero or bit one, so each bit is naturally spread over the whole image feature space. Moreover, bits act only as selection signals and every candidate feature is directly generated from the cover image features \(Z_{\mathrm{i}}\). As a result, \(Z_{\mathrm{m}}\) remains tightly coupled to the cover image features, jointly addressing the locality and weak-coupling issues of prior message processing blocks.

\subsection{Loss Function}
\textbf{Image Loss}
During the encoding stage, the encoder takes the cover image \(I_{\mathrm{c}}\) and the watermark \(M\) as inputs and outputs the watermarked image \(I_{\mathrm{w}}\). To keep the visual quality of \(I_{\mathrm{w}}\) close to \(I_{\mathrm{c}}\), the image loss is defined as follows:
\begin{equation}
\mathcal{L}_{\mathrm{Image}} = \mathcal{L}_{\mathrm{MSE}}\!\left(I_{\mathrm{c}},\, I_{\mathrm{w}}\right).
\end{equation}
\textbf{Message Loss}
During the decoding stage, the decoder takes the distorted image \(I_{\mathrm{d}}=\mathcal{N}(I_{\mathrm{w}})\) as input and predicts the extracted watermark \(M_{\mathrm{ex}}\).
To ensure robust and accurate decoding, the message loss is defined as follows:
\begin{equation}
\mathcal{L}_{\mathrm{Message}}
\;=\; \mathcal{L}_{\mathrm{MSE}}\!\left(M,\, M_{\mathrm{ex}}\right).
\end{equation}
\textbf{Total Loss}
Visual quality and decoding accuracy present a trade off. The total loss is a weighted sum of the two losses:
\begin{equation}
\mathcal{L}_{\mathrm{Total}}
=
\lambda_{1}\,\mathcal{L}_{\mathrm{Image}}
+
\lambda_{2}\,\mathcal{L}_{\mathrm{Message}},
\end{equation}
where \(\lambda_{1}\) and \(\lambda_{2}\) balance the trade off between visual quality and decoding accuracy.

\begin{figure*}[!t]
\centering
\includegraphics[width=0.90\textwidth]{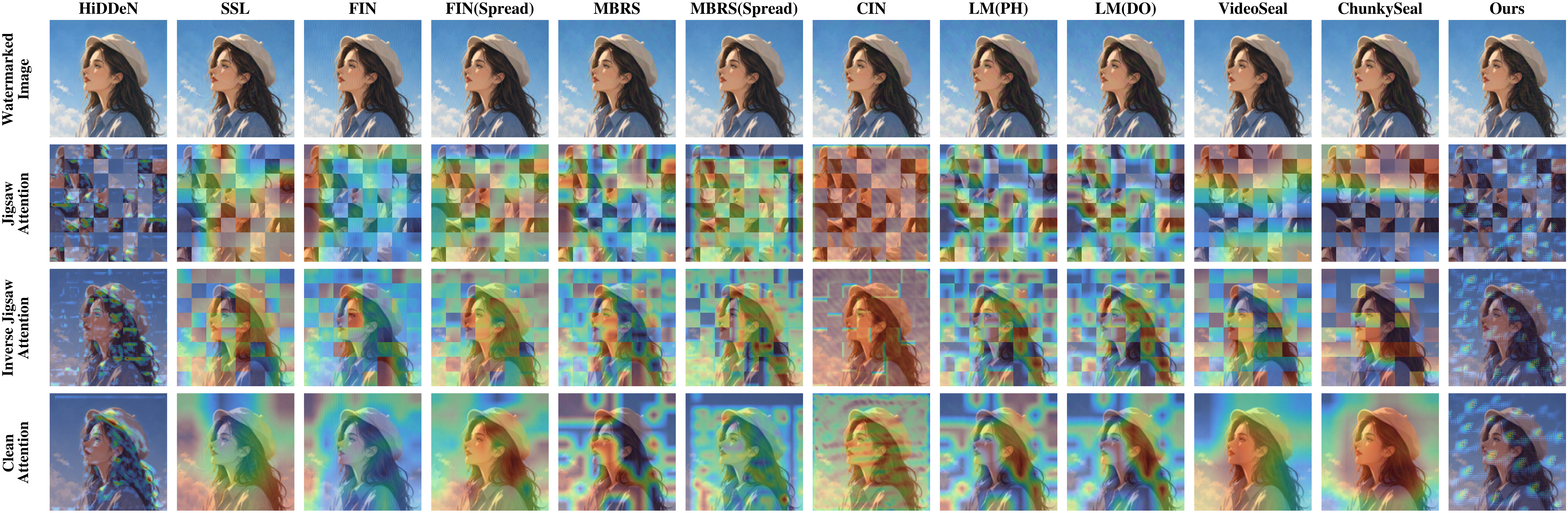}
\caption[Decoder attention under jigsaw distortion]{Decoder attention under jigsaw distortion. Rows from top to bottom show the watermarked image, the attention map on the jigsaw distorted image, the inverse-permuted version of that attention map, and the attention map on the undistorted image. Closer agreement between the third and fourth rows indicates better spatial realignment.}
\label{Disorder}
\end{figure*}

\section{Experimental Results}
Due to space limitations, detailed experimental settings and experiments for cover images of different shapes are provided in the Supplementary Material, in Section~\emph{Experimental Settings} and Table~\ref{tab:ours_multishape_elastic_shapeaware} of Section~\emph{Cover Images with Different Shapes}, respectively.

\begin{table}[t]
\centering
{\footnotesize
\setlength{\tabcolsep}{3pt}
\renewcommand{\arraystretch}{1.00}
\begin{tabular}{@{}lccc@{}}
\toprule
\textbf{Method} & \textbf{PSNR (dB)}$\uparrow$ & \textbf{SSIM}$\uparrow$ & \textbf{LPIPS}$\downarrow$ \\
\midrule
HiDDeN & 33.59 & 0.9654 & 0.0095 \\
SSL & 35.98 & 0.9547 & 0.0264 \\
FIN & 37.12 & 0.9393 & 0.0567 \\
FIN(Spread) & 37.01 & 0.9377 & 0.0130 \\
MBRS & 36.30 & 0.9585 & 0.0173 \\
MBRS(Spread) & 37.08 & 0.9750 & 0.0093 \\
CIN & 36.39 & 0.9639 & 0.0115 \\
LM(PH) & 36.71 & 0.9620 & 0.0260 \\
LM(DO) & 36.78 & 0.9670 & 0.0241 \\
VideoSeal & 36.09 & 0.9628 & 0.0113 \\
ChunkySeal & 36.87 & 0.9749 & 0.0132 \\
\rowcolor{gray!20}
Ours & 40.82 & 0.9779 & 0.0074 \\
\bottomrule
\end{tabular}
}
\caption{Visual quality of different methods.}
\label{tab:visual_quality}
\end{table}

\begin{table*}[t]
\centering
{\footnotesize
\setlength{\tabcolsep}{4.2pt}
\renewcommand{\arraystretch}{1.12}
\resizebox{\textwidth}{!}{%
\begin{tabular}{c|cccccc|cccccc|cccc|c}
\toprule
\multirow{2}{*}{\textbf{Method}} &
\multicolumn{6}{c|}{\textbf{Signal Distortions}} &
\multicolumn{6}{c|}{\textbf{Geometric Transformations}} &
\multicolumn{4}{c|}{\textbf{Photometric Transformations}} &
\multirow{2}{*}{\shortstack{\textbf{AVG}\\(\%)}} \\
&
\shortstack{\textbf{JPEG}\\(\%)} & \shortstack{\textbf{MF}\\(\%)} & \shortstack{\textbf{GF}\\(\%)} & \shortstack{\textbf{DP}\\(\%)} & \shortstack{\textbf{S\&P}\\(\%)} & \shortstack{\textbf{GN}\\(\%)} & \shortstack{\textbf{Erase}\\(\%)} &
\shortstack{\textbf{C\&R}\\(\%)} & \shortstack{\textbf{Shear}\\(\%)} & \shortstack{\textbf{Rotate}\\(\%)} & \shortstack{\textbf{Elastic}\\(\%)} & \shortstack{\textbf{Jigsaw}\\(\%)} &
\shortstack{\textbf{Hue}\\(\%)} & \shortstack{\textbf{Bright}\\(\%)} & \shortstack{\textbf{Contrast}\\(\%)} & \shortstack{\textbf{Saturate}\\(\%)} &
\\
\midrule
HiDDeN & 54.34 & 56.51 & 57.20 & 57.55 & 54.34 & 60.16 & 55.56 & 59.03 & 54.47 & 58.38 & 63.80 & 59.90 & 60.16 & 56.68 & 60.55 & 62.02 & 58.17 \\
SSL & 85.42 & 100.00 & 98.44 & 94.88 & 58.33 & 89.76 & 55.21 & 65.97 & 67.88 & 71.88 & 85.42 & 54.69 & 100.00 & 99.78 & 99.96 & 99.83 & 82.96 \\
FIN & 98.87 & 88.11 & 78.99 & 100.00 & 87.67 & 100.00 & 67.80 & 47.92 & 49.44 & 49.52 & 46.96 & 50.09 & 100.00 & 98.57 & 99.52 & 100.00 & 78.97 \\
FIN(Spread) & 85.07 & 69.27 & 69.10 & 96.61 & 79.95 & 97.22 & 70.49 & 49.22 & 48.52 & 49.18 & 47.92 & 52.08 & 97.40 & 96.09 & 96.53 & 97.53 & 75.14 \\
MBRS & 98.52 & 94.70 & 96.70 & 99.83 & 99.91 & 99.83 & 83.07 & 52.34 & 53.12 & 50.78 & 89.32 & 47.83 & 100.00 & 98.96 & 99.74 & 100.00 & 85.29 \\
MBRS(Spread) & 51.04 & 53.91 & 63.45 & 96.53 & 89.84 & 68.40 & 77.95 & 47.48 & 53.56 & 47.44 & 59.98 & 50.78 & 98.65 & 96.44 & 98.35 & 99.39 & 72.08 \\
CIN & 75.35 & 96.96 & 98.87 & 99.57 & 99.91 & 100.00 & 99.22 & 54.25 & 94.05 & 89.97 & 94.44 & 53.47 & 100.00 & 99.39 & 99.96 & 100.00 & 90.96 \\
LM(PH) & 97.83 & 96.79 & 98.96 & 97.05 & 99.65 & 99.83 & 69.44 & 50.52 & 52.08 & 49.52 & 81.16 & 49.65 & 98.83 & 97.05 & 99.13 & 99.74 & 83.58 \\
LM(DO) & 96.79 & 96.61 & 98.44 & 96.44 & 99.05 & 99.31 & 66.93 & 49.05 & 57.55 & 49.91 & 88.02 & 49.91 & 97.74 & 96.70 & 98.57 & 99.13 & 83.76 \\
VideoSeal & 99.91 & 98.78 & 99.48 & 99.91 & 100.00 & 99.65 & 98.35 & 51.48 & 98.44 & 97.09 & 99.74 & 58.77 & 100.00 & 99.74 & 100.00 & 100.00 & 93.83 \\
ChunkySeal & 91.32 & 100.00 & 100.00 & 99.83 & 99.91 & 100.00 & 97.48 & 82.29 & 94.70 & 87.85 & 99.91 & 58.25 & 100.00 & 99.87 & 99.91 & 100.00 & 94.46 \\
\midrule
\rowcolor{gray!20}
Ours & 96.70 & 99.39 & 99.74 & 100.00 & 100.00 & 98.96 & 98.44 & 98.44 & 99.52 & 99.48 & 99.91 & 100.00 & 100.00 & 99.87 & 100.00 & 100.00 & 99.40 \\
\bottomrule
\end{tabular}%
}
}
\caption{Benchmark comparisons on robustness against diverse distortions.}
\label{tab:methods_by_distortions}
\end{table*}

\begin{table}[t]
\centering
{\footnotesize
\setlength{\tabcolsep}{1.4pt}
\renewcommand{\arraystretch}{0.96}
\begin{tabular}{@{}l*{6}{c}@{}}
\toprule
{\scriptsize\textbf{Method}} & {\scriptsize\shortstack{\textbf{Crayon}\\(\%)}} & {\scriptsize\shortstack{\textbf{Film}\\(\%)}} & {\scriptsize\shortstack{\textbf{Heavy}\\(\%)}} & {\scriptsize\shortstack{\textbf{Layering}\\(\%)}} & {\scriptsize\shortstack{\textbf{Sketch}\\(\%)}} & {\scriptsize\shortstack{\textbf{AVG}\\(\%)}} \\
\midrule
HiDDeN      & 59.55 & 55.21 & 52.95 & 60.68 & 56.51 & 56.98 \\
SSL         & 66.15 & 62.59 & 59.81 & 66.58 & 59.64 & 62.95 \\
FIN         & 98.31 & 97.40 & 99.48 & 98.96 & 96.88 & 98.21 \\
FIN(Spread) & 95.75 & 93.23 & 96.44 & 96.35 & 93.58 & 95.07 \\
MBRS        & 98.35 & 93.84 & 99.39 & 100.00 & 99.14 & 98.14 \\
MBRS(Spread)& 90.36 & 81.86 & 91.15 & 86.81 & 82.47 & 86.53 \\
CIN         & 98.31 & 89.15 & 95.92 & 98.61 & 97.74 & 95.95 \\
LM(PH)      & 97.57 & 87.50 & 92.71 & 96.70 & 96.18 & 94.13 \\
LM(DO)      & 97.66 & 90.36 & 85.07 & 93.49 & 94.88 & 92.29 \\
VideoSeal   & 98.05 & 96.96 & 99.31 & 100.00 & 89.76 & 96.82 \\
ChunkySeal  & 98.15 & 96.44 & 97.83 & 99.83 & 99.27 & 98.30 \\
\rowcolor{gray!20}
Ours        & 98.44 & 99.83 & 99.91 & 100.00 & 99.57 & 99.55 \\
\bottomrule
\end{tabular}
}
\caption{Benchmark comparisons on robustness against diverse black-box distortions.}
\label{tab:blackbox_distortion}
\end{table}

\begin{table*}[t]
\centering
{\scriptsize
\setlength{\tabcolsep}{3.2pt}
\renewcommand{\arraystretch}{0.98}
\resizebox{\textwidth}{!}{%
\begin{tabular}{c|c|cccccc|cccccc|cccc|c}
\toprule
\textbf{Method} &
\shortstack{\textbf{PSNR}\\\textbf{(dB)}} &
\shortstack{\textbf{JPEG}\\(\%)} & \shortstack{\textbf{MF}\\(\%)} & \shortstack{\textbf{GF}\\(\%)} & \shortstack{\textbf{DP}\\(\%)} & \shortstack{\textbf{S\&P}\\(\%)} & \shortstack{\textbf{GN}\\(\%)} &
\shortstack{\textbf{Erase}\\(\%)} & \shortstack{\textbf{C\&R}\\(\%)} & \shortstack{\textbf{Shear}\\(\%)} & \shortstack{\textbf{Rotate}\\(\%)} &
\shortstack{\textbf{Elastic}\\(\%)} & \shortstack{\textbf{Jigsaw}\\(\%)} &
\shortstack{\textbf{Hue}\\(\%)} & \shortstack{\textbf{Bright}\\(\%)} & \shortstack{\textbf{Contrast}\\(\%)} & \shortstack{\textbf{Saturate}\\(\%)} &
\shortstack{\textbf{AVG}\\(\%)} \\
\midrule
w/o CAS & 38.85 & 50.43 & 71.53 & 77.34 & 98.61 & 98.00 & 96.09 & 82.99 & 49.31 & 49.83 & 50.17 & 51.13 & 52.43 & 99.31 & 98.13 & 99.09 & 98.96 & 76.46 \\
w/o IAL & 39.96 & 51.04 & 82.12 & 84.55 & 98.96 & 97.31 & 95.40 & 80.56 & 50.26 & 50.00 & 50.74 & 50.43 & 91.93 & 99.13 & 98.61 & 99.39 & 99.09 & 79.97 \\
\rowcolor{gray!20}
Ours & 40.82 & 96.70 & 99.39 & 99.74 & 100.00 & 100.00 & 98.96 & 98.44 & 98.44 & 99.52 & 99.48 & 99.91 & 100.00 & 100.00 & 99.87 & 100.00 & 100.00 & 99.40 \\
\bottomrule
\end{tabular}%
}
}
\caption{Ablation on visual quality and robustness under diverse distortions.}
\label{tab:abla-gd-combined}
\end{table*}

\subsection{Analysis of Encoders}
The Motivation Section argues that prior MPBs suffer either from insufficient spatial spreading or from weak coupling to image content, which limits visual quality.
Figure~\ref{InfoLoss} supports this claim.

Methods following MPB(a), including FIN~\citep{fang2023flow}, MBRS~\citep{jia2021mbrs}, LM(PH), and LM(DO)~\citep{qiu2025lightweight}, directly upsample the message and fail to distribute each bit across the full image.
Methods following MPB(b), including MBRS (Spread)~\citep{jia2021mbrs}, FIN (Spread)~\citep{fang2023flow}, and CIN~\citep{ma2022towards}, achieve broader dispersion by projecting the message before upsampling.
However, their residual maps remain nearly unchanged across different cover images, indicating limited content adaptivity and helping explain their weaker fidelity.

Methods following MPB(c), including HiDDeN~\citep{zhu2018hidden}, VideoSeal~\citep{fernandez2024video}, and ChunkySeal~\citep{petrov2025we}, achieve full-image spreading by broadcasting the message.
However, the redundant message feature imposes a heavy visual burden and introduces excessive residuals in smooth, perceptually sensitive background regions.
By contrast, CASIAL produces image-wide residuals whose patterns vary with the cover image while avoiding smooth sensitive regions.
This demonstrates both broad spreading and strong content coupling with high visual quality, which is precisely the behavior that CAS is designed to induce.

\subsection{Analysis of Decoders}

When geometric transformations displace local structure, a robust decoder must aggregate information globally and redirect attention toward reliable regions. 
Figure~\ref{Disorder} visualizes decoder attention under jigsaw distortion. For each method, we show the watermarked image, the attention map on the distorted image, the inverse-permuted version of that attention map, and the attention map on the clean image. Successful decoding requires the inverse-permuted attention to align with the clean attention. For the baselines, the discrepancy between these two maps is substantial, indicating poor spatial realignment after permutation. Our decoder, in contrast, produces an inverse-permuted attention map that closely matches the clean one, which suggests that IAL is effectively recovering the relevant spatial evidence under severe desynchronization.

\subsection{Visual Quality}
Table~\ref{tab:visual_quality} compares visual quality using PSNR, SSIM, and LPIPS. CASIAL achieves the best performance on all three metrics, with \(40.82\) dB PSNR, \(0.9779\) SSIM, and \(0.0074\) LPIPS. It improves PSNR by \(3.70\) dB over FIN and also outperforms MBRS (Spread) in both SSIM and LPIPS. These results show that CASIAL improves robustness without introducing stronger visible artifacts. Its cover image-aware message spreading and JND-based attenuation preserve high visual fidelity while maintaining reliable watermark decoding.

\subsection{White-box Robustness}
We evaluate white-box robustness against six geometric, six signal, and four photometric distortions (Table~\ref{tab:methods_by_distortions}); visual examples are provided in Supplementary Fig.~\ref{Distortion}. Erasing removes \(80\%\) of the image, crop \& resize retains \(20\%\), shear and rotation are averaged over \(\pm60^\circ\) and \(\pm45^\circ\), elastic deformation uses \(\alpha=2.0\), and jigsaw uses an \(8{\times}8\) grid. CASIAL achieves at least \(98.44\%\) accuracy across all geometric transformations, demonstrating robustness to both region removal and spatial desynchronization. This supports the complementary roles of CAS in distributing watermark evidence and IAL in aggregating displaced evidence.

CASIAL also achieves at least \(96.70\%\) accuracy under all signal distortions and \(99.87\%\) under all photometric transformations. Overall, it obtains the highest average accuracy of \(99.40\%\) across all \(16\) distortions, outperforming ChunkySeal by \(4.94\) percentage points while maintaining high visual quality.

\subsection{Black-box Robustness}

Visual examples are shown in Supplementary Fig.~\ref{Black}.
Table~\ref{tab:blackbox_distortion} reports robustness against five unseen black-box distortions. CASIAL achieves the highest average accuracy of \(99.55\%\), outperforming ChunkySeal by \(1.25\) percentage points. It obtains the best or tied-best result on every distortion, with \(98.44\%\) on Crayon, \(99.83\%\) on Film, \(99.91\%\) on Heavy, \(100.00\%\) on Layering, and \(99.57\%\) on Sketch. Although several baselines perform well on individual distortions, their accuracy varies more substantially across styles.

These results demonstrate that CASIAL generalizes beyond the white-box distortions used during training. The consistently high accuracy suggests that CAS and IAL learn spatially distributed and content-adaptive watermark representations that remain decodable under unseen appearance changes.

\section{Ablation Study}
Due to space limitations, additional experiments are provided in the Supplementary Material, including robustness evaluations under varying geometric-distortion strengths (Fig.~\ref{fig:geometric_qualitative}) and analyses of the effects of JND on perceptual quality and residual placement (Table~\ref{tab:jnd_ablation_noclip} and Fig.~\ref{JND}).

\subsection{Importance of CAS and IAL}
Table~\ref{tab:abla-gd-combined} evaluates the contributions of CAS and IAL. Removing CAS reduces the average decoding accuracy from \(99.40\%\) to \(76.46\%\) and PSNR from \(40.82\) dB to \(38.85\) dB. The variant also falls to near chance level under crop \& resize, shear, rotation, elastic deformation, and jigsaw permutation. This shows that CAS improves both geometric robustness and visual quality by distributing cover-conditioned watermark evidence across the image.

Removing IAL reduces the average accuracy to \(79.97\%\), while retaining a relatively high PSNR of \(39.96\) dB. Its poor performance under most geometric transformations indicates that, although the watermark can still be embedded imperceptibly, the decoder cannot reliably aggregate spatially displaced evidence.

These results confirm that CAS and IAL are complementary: CAS controls how watermark evidence is distributed, while IAL enables its aggregation and realignment after geometric transformations. Together, they address region removal and spatial desynchronization.

\section{Conclusion}
We presented CASIAL, a geometry-robust image watermarking framework. To address region removal and desynchronization, CAS distributes cover-aware message evidence across the image, while IAL aggregates displaced evidence under spatial misalignment.
Experiments against 11 state-of-the-art baselines show that CASIAL achieves the highest average white-box robustness while preserving strong visual quality. It also generalizes to unseen black-box distortions and diverse image shapes. Ablation studies verify the complementary roles of CAS in distributed encoding and IAL in robust decoding. These results suggest that geometric robustness requires not only distortion augmentation, but also effective message spreading and spatial alignment.

\bibliography{refs}

\end{document}


\maketitle

\section{CAS Strategy}
An overview of the CAS strategy is shown in Fig.~\ref{MessageProcessBlockFig}(d), and the algorithmic flow is presented in Algorithm~\ref{alg:spcame}.

\begin{algorithm}[H]
\footnotesize
\caption{CAS: Cover Image-Aware Message Spreading}
\label{alg:spcame}
\begin{algorithmic}[1]
\Require Cover-image feature $Z_{\mathrm{i}}\!\in\!\mathbb{R}^{C\times \tfrac{H}{4}\times \tfrac{W}{4}}$; binary message $M=[m_{1},\ldots,m_{L}]\in\{0,1\}^{L}$
\Ensure Message feature $Z_{\mathrm{m}}\!\in\!\mathbb{R}^{C\times \tfrac{H}{4}\times \tfrac{W}{4}}$
\Statex
\State \textbf{Candidate feature generation}
\State $F^{(0)} \gets \mathrm{IAL}_{0}(Z_{\mathrm{i}})$ \Comment{candidate for bit 0}
\State $F^{(1)} \gets \mathrm{IAL}_{1}(Z_{\mathrm{i}})$ \Comment{candidate for bit 1}
\Statex
\State \textbf{Binary gated feature selection (BGFS)}
\For{$k = 1$ to $L$}
  \State $S_{k} \gets (1-m_{k})\,F^{(0)} + m_{k}\,F^{(1)}$ \Comment{$S_k\!\in\!\mathbb{R}^{C\times \tfrac{H}{4}\times \tfrac{W}{4}}$}
\EndFor
\State $\tilde{S} \gets \operatorname{Concat}_{\text{channel}}(S_{1},\ldots,S_{L})$ \Comment{$\tilde{S}\!\in\!\mathbb{R}^{(L\cdot C)\times \tfrac{H}{4}\times \tfrac{W}{4}}$}
\Statex
\State \textbf{Attention-based feature fusion}
\State $Z_{\mathrm{m}} \gets \mathrm{IAL}_{2}(\tilde{S})$
\State \Return $Z_{\mathrm{m}}$
\end{algorithmic}
\end{algorithm}
\section{Experimental Settings}
\begin{figure*}[!t]
\centering
\includegraphics[width=0.92\textwidth]{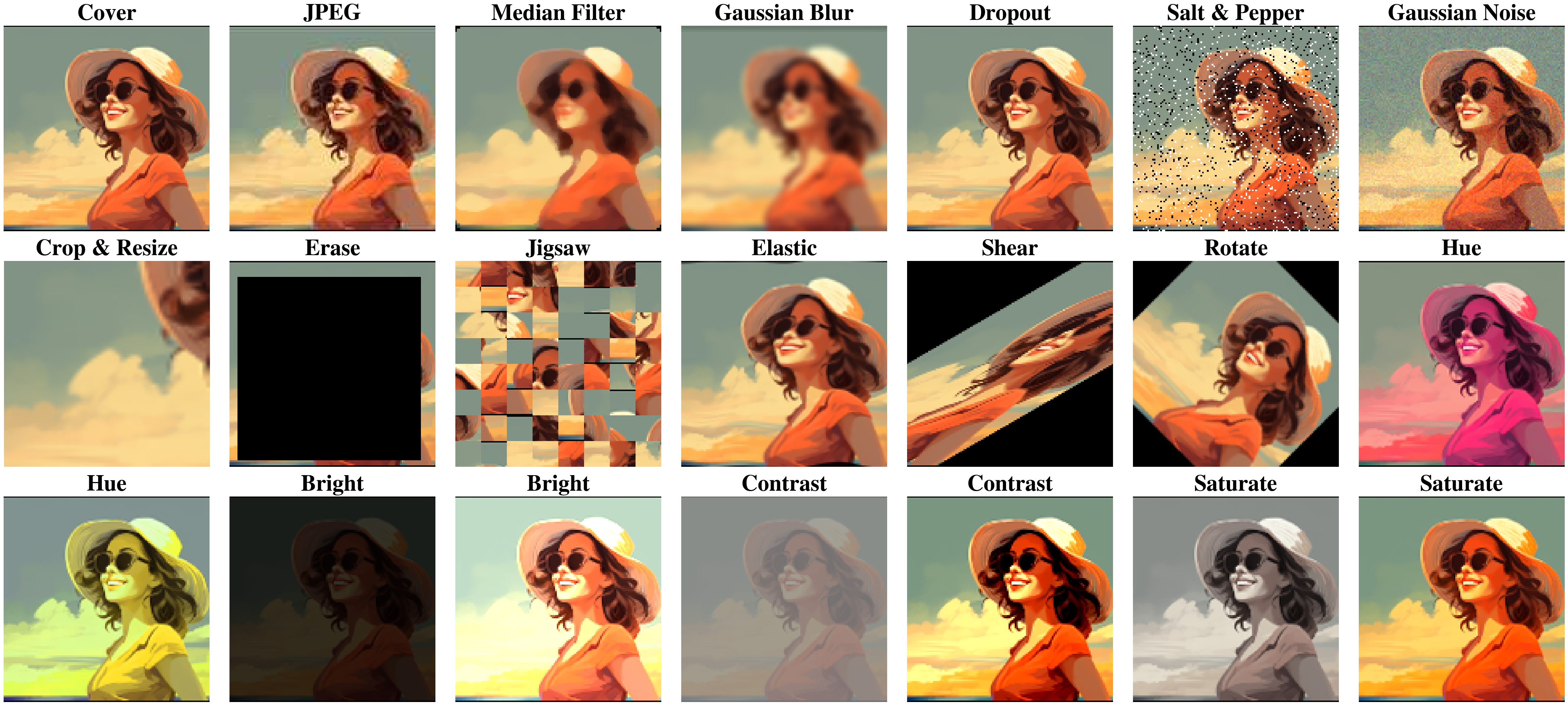}
\caption[Examples of distortions]{Examples of distortions considered in our experiments.
From left to right and top to bottom, the figure shows the cover image;
six signal distortions, including JPEG compression, median filtering, Gaussian filtering, dropout, salt-and-pepper noise, and Gaussian noise;
six geometric transformations, including crop \& resize, erasing, jigsaw permutation, elastic deformation, shear, and rotation;
and eight photometric transform examples, where hue, brightness, contrast, and saturation adjustments are each shown in two directions.}
\label{Distortion}
\end{figure*}

\textbf{Dataset and Settings}
We train on MS-COCO~\citep{lin2014microsoft} and evaluate it on USC-SIPI~\citep{USC_SIPI_Image_Database}. Unless otherwise specified, images are resized to \(128\times128\), the message length is \(L=64\), and \(\lambda_1=20\), \(\lambda_2=1\). We use Adam~\citep{kingma2014adam} with a learning rate of \(1\times10^{-5}\) and default settings. All experiments are implemented in PyTorch~\citep{paszke2019pytorch} and run on two NVIDIA RTX A40 GPUs.
\\
\textbf{Benchmarks}
To comprehensively assess the robustness and visual quality of our method, we compare CASIAL with 11 state-of-the-art baselines. These include encoder-noise-layer-decoder (END)-based methods, namely HiDDeN \citep{zhu2018hidden}, MBRS and MBRS with a spreading module \citep{jia2021mbrs}, Lightweight Mark in its DO and PH variants \citep{qiu2025lightweight}, VideoSeal \citep{fernandez2024video}, and ChunkySeal \citep{petrov2025we}; invertible-block-based methods, namely CIN \citep{ma2022towards}, FIN and FIN with a spreading module \citep{fang2023flow}; and the iterative optimization method SSL \citep{fernandez2022watermarking}. All baselines are retrained from the official implementations released by their authors. For fair comparison, all models are trained under the same noise settings, image resolution, and message length.\\
\textbf{Distortions}
To thoroughly evaluate robustness under various distortions, we consider three groups of distortions: six geometric transformations, namely crop \& resize (C\&R), erasing, jigsaw permutation, elastic deformation, shear, and rotation, whose visual effects are shown in Fig.~\ref{Distortion}; six signal distortions, namely JPEG compression, median filtering (MF), Gaussian filtering (GF), dropout (DP), salt-and-pepper noise (S\&P), and Gaussian noise (GN); and four photometric transformations, namely hue, brightness, contrast, and saturation adjustments. During training, we adopt differentiable implementations based on Kornia \citep{riba2020kornia} for these white-box distortions, which allows gradients to be back-propagated through the distortion layer. During testing, we use practical implementations of the same distortions.

\begin{figure}[!t]
\centering
\includegraphics[width=\columnwidth]{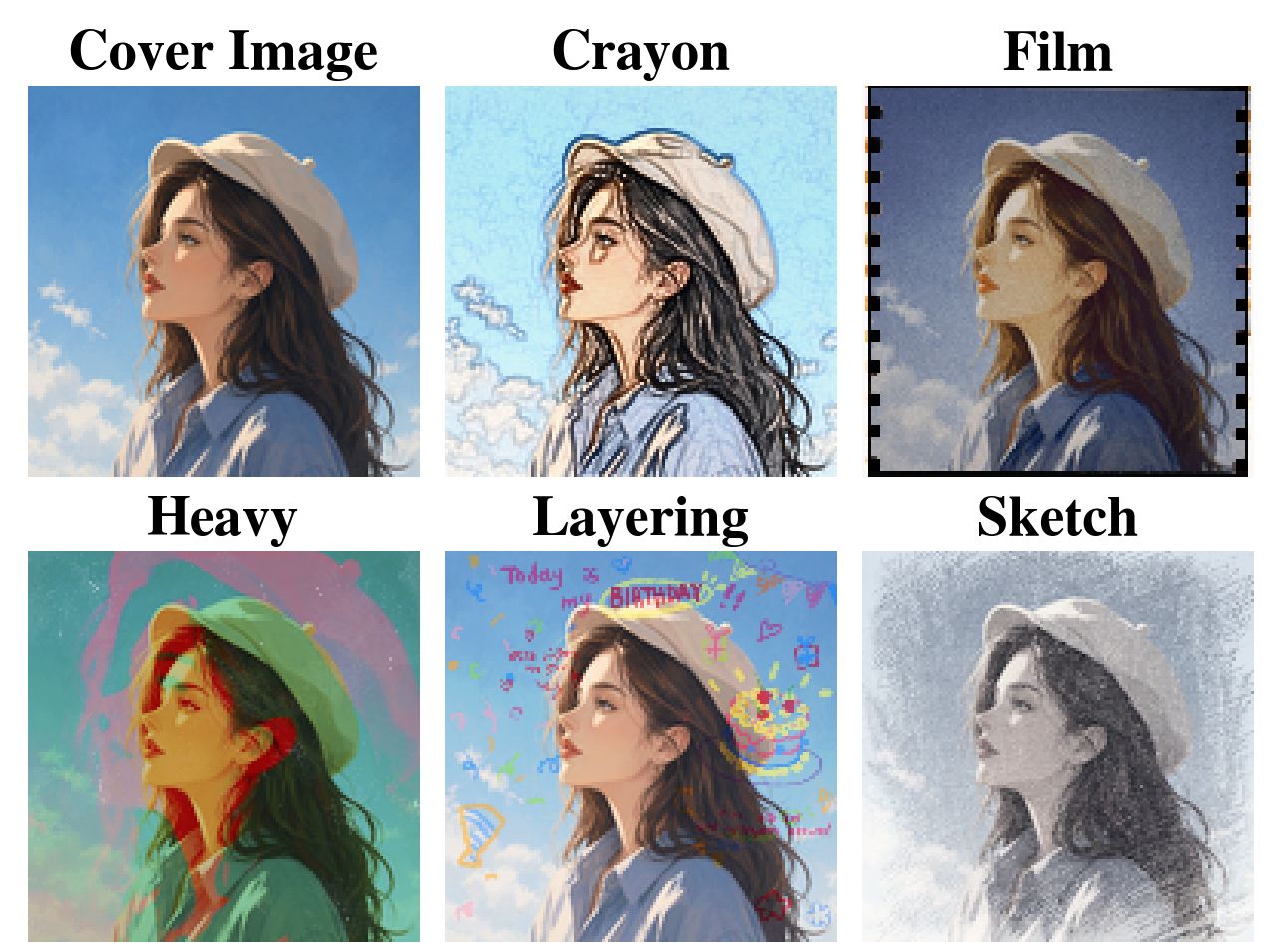}
\caption[Examples of black-box image distortions]{Examples of black-box image distortions used for robustness evaluation. From left to right, we show the original cover image and five black-box distortions: Crayon, Film, Heavy, Layering, and Sketch.}
\label{Black}
\end{figure}

In addition to the trainable white-box distortions, we also evaluate transfer robustness against five black-box distortions that cannot be inserted into the training pipeline, namely Crayon, Film, Heavy, Layering, and Sketch, as illustrated in Fig.~\ref{Black}.\\
\textbf{Evaluation metrics}
Visual quality is evaluated using peak signal-to-noise ratio (PSNR), structural similarity index (SSIM)~\citep{wang2004image}, and learned perceptual image patch similarity (LPIPS)~\citep{zhang2018unreasonable}. Robustness is measured by decoding bit accuracy (ACC) under each distortion.

\section{Cover Images with Different Shapes}
Since CAS derives the message feature from the cover image feature, the encoded message representation is naturally tied to the spatial shape of the input cover image.
This allows CASIAL to handle cover images with different resolutions and aspect ratios without changing the model architecture or adding any external shape conversion module.
In contrast, message-only processing blocks that generate a fixed spatial message feature are less flexible when the input shape changes.

\begin{table*}[!t]
\centering
{\scriptsize
\setlength{\tabcolsep}{3.2pt}
\renewcommand{\arraystretch}{0.98}
\resizebox{\textwidth}{!}{%
\begin{tabular}{c|cccccc|cccccc|cccc|c}
\toprule
\textbf{Shape} &
\shortstack{\textbf{JPEG}\\(\%)} & \shortstack{\textbf{MF}\\(\%)} & \shortstack{\textbf{GF}\\(\%)} & \shortstack{\textbf{DP}\\(\%)} & \shortstack{\textbf{S\&P}\\(\%)} & \shortstack{\textbf{GN}\\(\%)} &
\shortstack{\textbf{Erase}\\(\%)} & \shortstack{\textbf{C\&R}\\(\%)} & \shortstack{\textbf{Shear}\\(\%)} & \shortstack{\textbf{Rotate}\\(\%)} &
\shortstack{\textbf{Elastic}\\(\%)} & \shortstack{\textbf{Jigsaw}\\(\%)} &
\shortstack{\textbf{Hue}\\(\%)} & \shortstack{\textbf{Bright}\\(\%)} & \shortstack{\textbf{Contrast}\\(\%)} & \shortstack{\textbf{Saturate}\\(\%)} &
\shortstack{\textbf{AVG}\\(\%)} \\
\midrule
$128{\times}128$  & 96.70 & 99.39 & 99.74 & 100.00 & 100.00 & 98.96 & 98.44 & 98.44 & 99.52 & 99.48 & 99.91 & 100.00 & 100.00 & 99.87 & 100.00 & 100.00 & 99.40 \\
\midrule
$256{\times}128$  & 99.22 & 100.00 & 100.00 & 100.00 & 100.00 & 99.83 & 100.00 & 99.91 & 96.48 & 99.39 & 100.00 & 100.00 & 100.00 & 100.00 & 100.00 & 100.00 & 99.68 \\
\midrule
$256{\times}256$  & 100.00 & 100.00 & 100.00 & 100.00 & 100.00 & 100.00 & 100.00 & 99.91 & 99.96 & 99.96 & 100.00 & 100.00 & 100.00 & 100.00 & 100.00 & 100.00 & 99.99 \\
\midrule
$512{\times}256$  & 100.00 & 100.00 & 100.00 & 100.00 & 100.00 & 100.00 & 100.00 & 100.00 & 99.39 & 99.83 & 100.00 & 100.00 & 100.00 & 100.00 & 100.00 & 100.00 & 99.95 \\
\midrule
$512{\times}512$  & 100.00 & 100.00 & 100.00 & 100.00 & 100.00 & 100.00 & 100.00 & 99.91 & 99.91 & 99.83 & 99.91 & 100.00 & 100.00 & 100.00 & 100.00 & 100.00 & 99.97 \\
\bottomrule
\end{tabular}%
}
}
\caption{Multi-shape evaluation of the final Ours model with shape-aware Elastic.}
\label{tab:ours_multishape_elastic_shapeaware}
\end{table*}

We further verify the robustness of CASIAL under different input shapes.
Table~\ref{tab:ours_multishape_elastic_shapeaware} reports results on five input shapes, ranging from \(128{\times}128\) to \(512{\times}512\), including non-square shapes such as \(256{\times}128\) and \(512{\times}256\).
Across all tested shapes, CASIAL remains highly robust, with average decoding accuracy ranging from \(99.40\%\) to \(99.99\%\).
The default \(128{\times}128\) setting already achieves \(99.40\%\) average accuracy, while larger inputs reach nearly perfect accuracy, including \(99.99\%\) on \(256{\times}256\), \(99.95\%\) on \(512{\times}256\), and \(99.97\%\) on \(512{\times}512\).
The non-square cases are particularly important because they change the spatial support and aspect ratio of the cover image.
The stable performance on \(256{\times}128\) and \(512{\times}256\) indicates that CASIAL does not depend on a fixed square message layout.

These results support the motivation of cover image-aware message spreading.
Because the message feature is generated from the cover feature itself, CASIAL can adapt the watermark representation to the spatial support of the input image.
This differs from message-only processing blocks that first map the message to a predetermined spatial pattern and then combine it with image features.
By conditioning message features on the cover image, CASIAL can preserve robust decoding across different resolutions and aspect ratios without modifying the encoder-decoder architecture.
\section{Ablation Study}
\subsection{Performance under Different Geometric Distortion Strengths}
\begin{figure*}[t!]
\centering
\includegraphics[width=\textwidth]{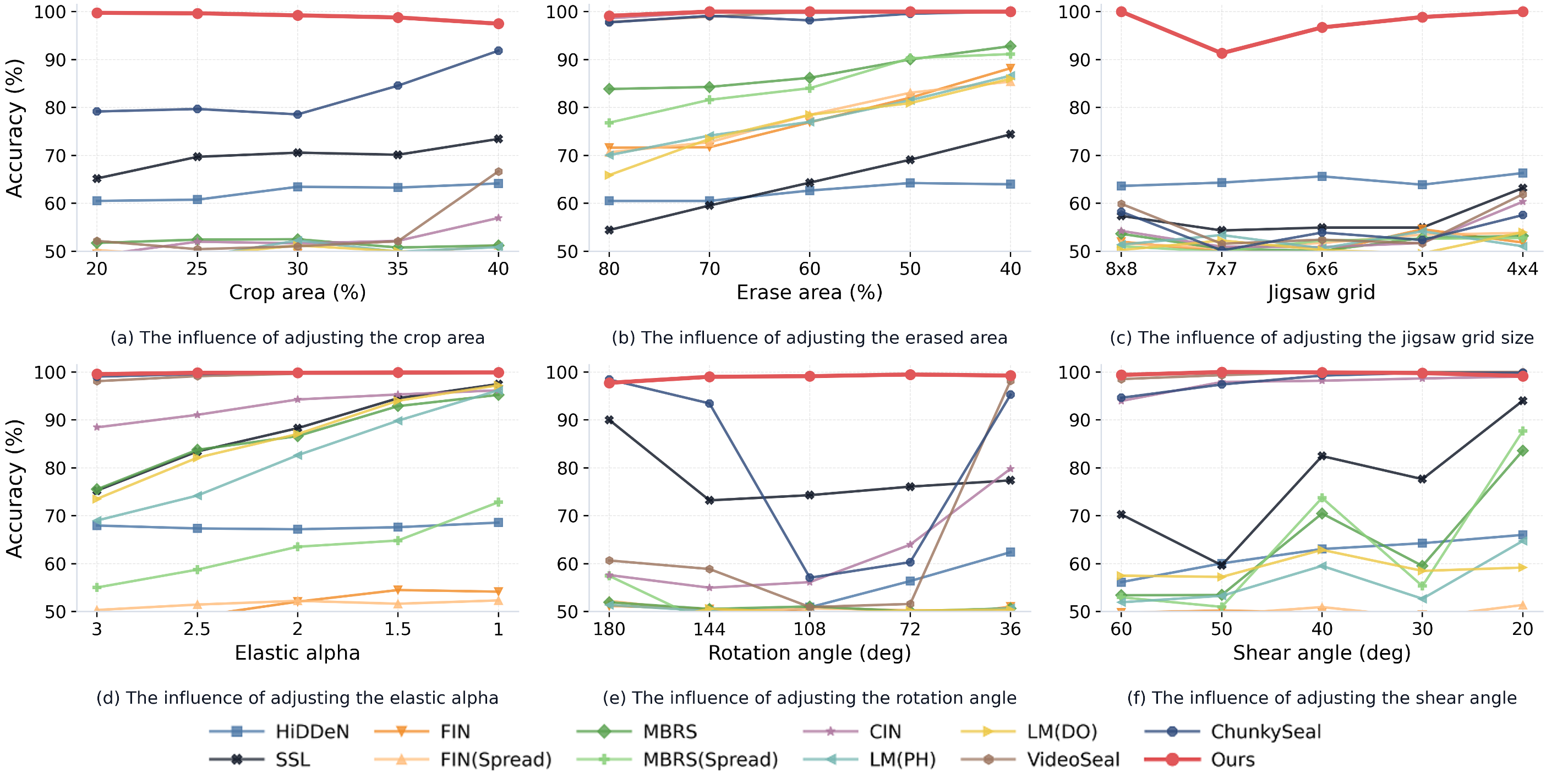}
\caption[Robustness under six representative geometric transformations]{Robustness under six representative geometric transformations.
Each subfigure reports decoding accuracy as the distortion strength varies.}
\label{fig:geometric_qualitative}
\end{figure*}

Figure~\ref{fig:geometric_qualitative} visualizes the decoding accuracy of 12 methods under six geometric transformations with varying strength levels.

For crop \& resize, the main challenge is that a large portion of the watermarked image can be removed before resizing.
CASIAL remains stable across a wide range of crop ratios, showing that the encoded message is not concentrated in a small local region.
This directly supports the role of CAS, since spatially distributed evidence makes decoding less dependent on any single image area.

For erasing, the distortion directly removes image content and creates missing regions.
CASIAL maintains strong decoding accuracy.
This indicates that the decoder can recover the message from the remaining visible regions, again suggesting that the watermark evidence is spatially distributed.

For jigsaw permutation, the image content is preserved but its spatial order is disrupted.
CASIAL shows strong robustness under different grid sizes, indicating that decoding does not rely only on fixed local spatial correspondence.
This highlights the role of IAL, since spatial attention helps aggregate displaced evidence after the local arrangement has been permuted.

For elastic deformation, the distortion introduces non-rigid local warping.
CASIAL remains robust as the deformation strength increases, suggesting that the encoded signal can tolerate smooth spatial displacement.
Because elastic deformation changes local geometry without fully destroying image content, successful decoding requires spatially tolerant feature aggregation rather than simple pixel-level invariance.

For rotation, the image undergoes global spatial transformation while preserving most visual content.
CASIAL maintains high decoding accuracy over varying rotation angles, showing that the learned representation is not strongly tied to the original upright coordinate system.
This further supports that the decoder can aggregate message evidence under global spatial misalignment.

For shear, the distortion produces strong directional geometric displacement.
CASIAL remains robust under both positive and negative shear angles, suggesting that the model can handle anisotropic spatial changes.
Together with the rotation and elastic results, this shows that CASIAL is effective not only for region removal, but also for severe spatial desynchronization.

\subsection{Importance of perceptual masking and JND attenuation}

\begin{table}[H]
\centering
{\footnotesize
\setlength{\tabcolsep}{3.5pt}
\renewcommand{\arraystretch}{1.02}
\begin{tabular}{@{}C{1.10cm}|*{4}{C{1.32cm}}@{}}
\toprule
\textbf{Method} & \shortstack{\textbf{PSNR}\\\textbf{(dB)}}$\uparrow$ & \textbf{SSIM}$\uparrow$ & \textbf{LPIPS}$\downarrow$ & \shortstack{\textbf{AVG}(\%)}\\
\midrule
w/o JND & 40.79 & 0.9673 & 0.0202 & 99.82 \\
\midrule
w/ JND  & 40.82 & 0.9779 & 0.0074 & 99.40 \\
\bottomrule
\end{tabular}
}
\caption{Effect of JND attenuation without hard PSNR clipping on COCO-test.}
\label{tab:jnd_ablation_noclip}
\end{table}

\begin{figure}[!t]
\centering
\includegraphics[width=\columnwidth]{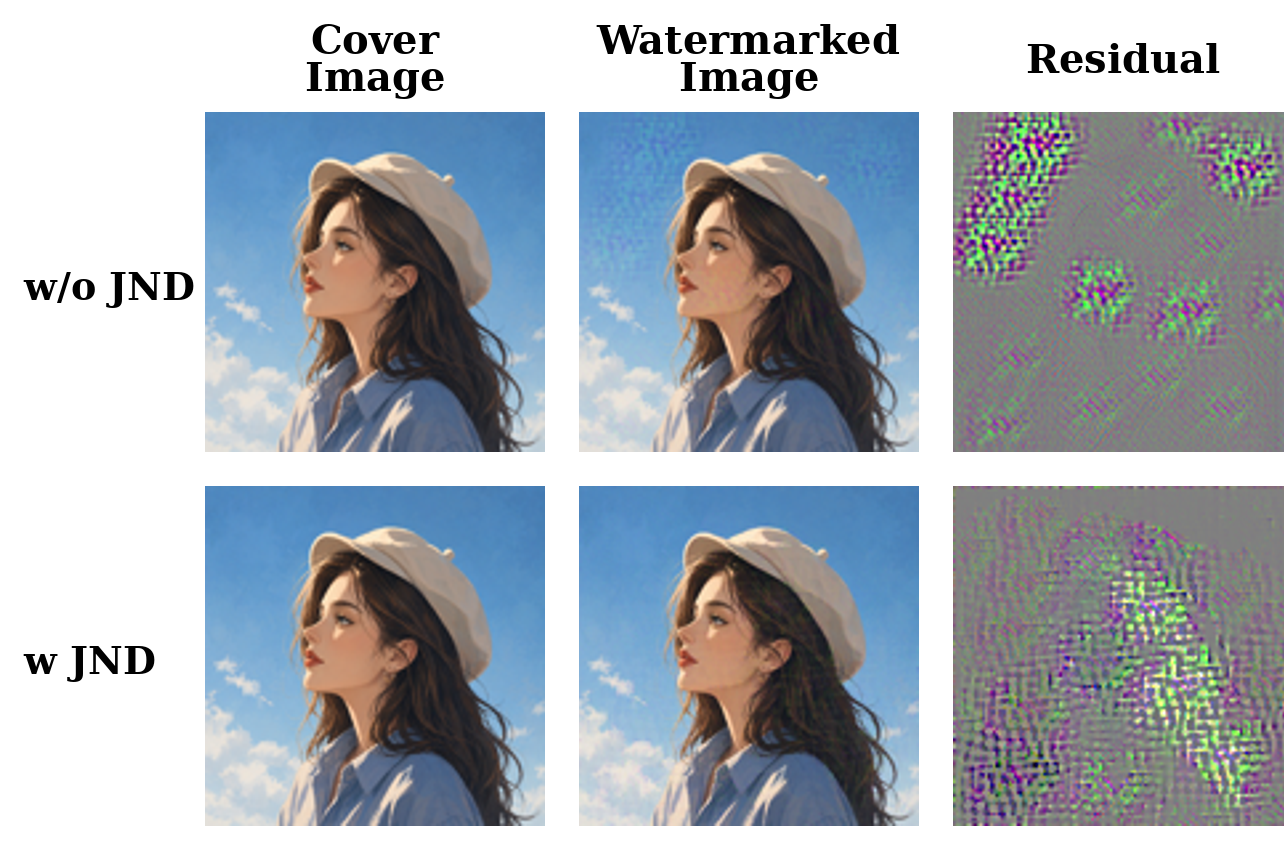}
\caption[Effect of JND-based residual attenuation]{Effect of JND-based residual attenuation. Without JND, the residual can concentrate in smooth background regions that are perceptually sensitive. With JND attenuation, the residual is suppressed in sensitive regions and shifted toward visually tolerant areas.}
\label{JND}
\end{figure}

Table~\ref{tab:jnd_ablation_noclip} evaluates the effect of JND-based residual attenuation.
The two variants have almost identical PSNR values, \(40.79\) dB without JND and \(40.82\) dB with JND, indicating that their overall perturbation budgets are comparable.
However, the perceptual metrics differ substantially.
Adding JND improves SSIM from \(0.9673\) to \(0.9779\) and reduces LPIPS from \(0.0202\) to \(0.0074\), showing a clear improvement in visual quality.
This confirms that similar pixel-level distortion can still lead to different perceptual quality.

Figure~\ref{JND} further explains this effect visually.
Without JND attenuation, the model tends to place strong residuals in smooth background regions, where distortions are more perceptible to humans.
With JND attenuation, the residual is suppressed in perceptually sensitive regions and preserved more in visually tolerant regions.
Thus, JND attenuation improves imperceptibility by changing where the watermark energy is placed, rather than simply reducing the overall distortion magnitude.
This also clarifies the role of JND in CASIAL: it improves perceptual residual placement, while the main robustness gains come from CAS and IAL.

\bibliography{refs}